# PixelVAE: A Latent Variable Model for Natural Images

**Ishaan Gulrajani**[1],[*] **Kundan Kumar**[1,2]**, Faruk Ahmed**[1]**, Adrien Ali Taiga**[1,3]**,**
**Francesco Visin**[1,5]**, David Vazquez**[1,4]**, Aaron Courville**[1,6]

[1] Montreal Institute for Learning Algorithms, Université de Montréal
[2] Department of Computer Science and Engineering, IIT Kanpur
[3] CentraleSupélec
[4] Computer Vision Center & Universitat Autonoma de Barcelona
[5] Politecnico di Milano
[6] CIFAR Fellow

## Abstract

Natural image modeling is a landmark challenge of unsupervised learning. Variational Autoencoders (VAEs) learn a useful latent representation and model global structure well but have difficulty capturing small details. PixelCNN models details very well, but lacks a latent code and is difficult to scale for capturing large structures. We present PixelVAE, a VAE model with an autoregressive decoder based on PixelCNN. Our model requires very few expensive autoregressive layers compared to PixelCNN and learns latent codes that are more compressed than a standard VAE while still capturing most non-trivial structure. Finally, we extend our model to a hierarchy of latent variables at different scales. Our model achieves state-of-the-art performance on binarized MNIST, competitive performance on $64 \times 64$ ImageNet, and high-quality samples on the LSUN bedrooms dataset.

## 1 Introduction

Building high-quality generative models of natural images has been a long standing challenge. Although recent work has made significant progress (Kingma & Welling, 2014; van den Oord et al., 2016a;b), we are still far from generating convincing, high-resolution natural images.

Many recent approaches to this problem are based on an efficient method for performing amortized, approximate inference in continuous stochastic latent variables: the variational autoencoder (VAE) (Kingma & Welling, 2014) jointly trains a top-down decoder generative neural network with a bottom-up encoder inference network. VAEs for images typically use rigid decoders that model the output pixels as conditionally independent given the latent variables. The resulting model learns a useful latent representation of the data and effectively models global structure in images, but has difficulty capturing small-scale features such as textures and sharp edges due to the conditional independence of the output pixels, which significantly hurts both log-likelihood and quality of generated samples compared to other models.

PixelCNNs (van den Oord et al., 2016a;b) are another state-of-the-art image model. Unlike VAEs, PixelCNNs model image densities autoregressively, pixel-by-pixel. This allows it to capture fine details in images, as features such as edges can be precisely aligned. By leveraging carefully constructed masked convolutions (van den Oord et al., 2016b), PixelCNNs can be trained efficiently in parallel on GPUs. Nonetheless, PixelCNN models are still very computationally expensive. Unlike typical convolutional architectures they do not apply downsampling between layers, which means that each layer is computationally expensive and that the depth of a PixelCNN must grow linearly with the size of the images in order for it to capture dependencies between far-away pixels. PixelCNNs also do not explicitly learn a latent representation of the data, which can be useful for downstream tasks such as semi-supervised learning.

---

[*]Corresponding author; igul222@gmail.com

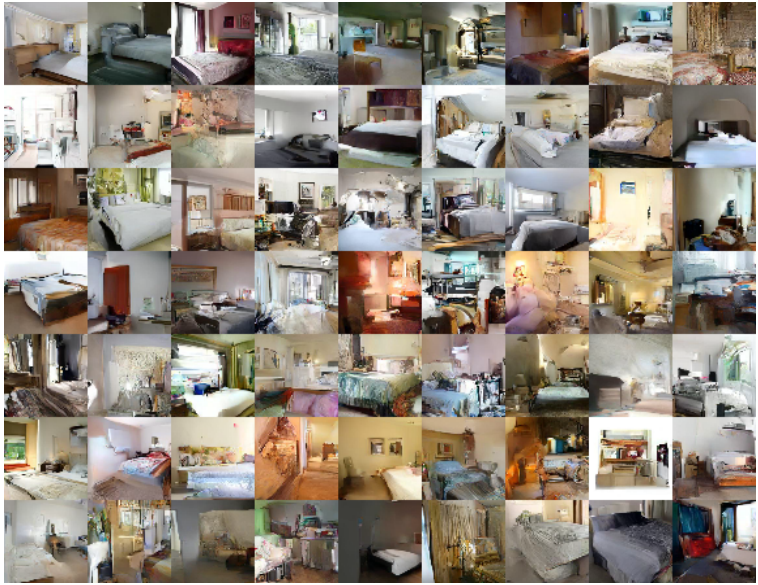

Figure 1: Samples from hierarchical PixelVAE on the LSUN bedrooms dataset.

Our contributions are as follows:

- We present PixelVAE, a latent variable model which combines the largely complementary advantages of VAEs and PixelCNNs by using PixelCNN-based masked convolutions in the conditional output distribution of a VAE.
- We extend PixelVAE to a hierarchical model with multiple stochastic layers and PixelCNN decoders at each layer. This lets us autoregressively model with PixelCNN not only the output pixels but also higher-level latent feature maps.
- On binarized MNIST, we show that PixelVAE: (1) achieves state-of-the-art performance, (2) can perform comparably to PixelCNN using far fewer computationally expensive autoregressive layers, and (3) can store less information in its latent variable than a standard VAE while still accounting for most non-trivial structure.
- We evaluate hierarchical PixelVAE on $64 \times 64$ ImageNet and the LSUN bedrooms dataset. On $64 \times 64$ ImageNet, we report competitive log-likelihood. On LSUN bedrooms, we generate high-quality samples and show that PixelVAE learns to model different properties of the scene with each of its multiple layers.

## 2 RELATED WORK

There have been many recent advancements in generative modeling of images. We briefly discuss some of these below, especially those that are related to our approach.

The Variational Autoencoder (VAE) (Kingma & Welling, 2014) is an elegant framework to train neural networks for generation and approximate inference jointly by optimizing a variational bound on the data log-likelihood. The use of normalizing flows (Rezende & Mohamed, 2015) improves the flexibility of the VAE approximate posterior. Based on this, Kingma et al. (2016) develop an efficient formulation of an autoregressive approximate posterior model using MADE (Germain et al., 2015). In our work, we avoid the need for such flexible inference models by using autoregressive priors.

Another promising recent approach is Generative Adversarial Networks (GANs) (Goodfellow et al., 2014), which pit a generator network and a discriminator network against each other. The generator tries to generate samples similar to the training data to fool the discriminator, and the discriminator tries to detect if the samples originate from the data distribution or not. Recent work has improved training stability (Radford et al., 2015; Salimans et al., 2016) and incorporated inference networks into the GAN framework (Dumoulin et al., 2016; Donahue et al., 2016). GANs generate compelling samples compared to our work, but still exhibit unstable training dynamics and are known to underfit by ignoring modes of the data distribution (Dumoulin et al., 2016). Further, it is difficult to accurately estimate the data likelihood in GANs.

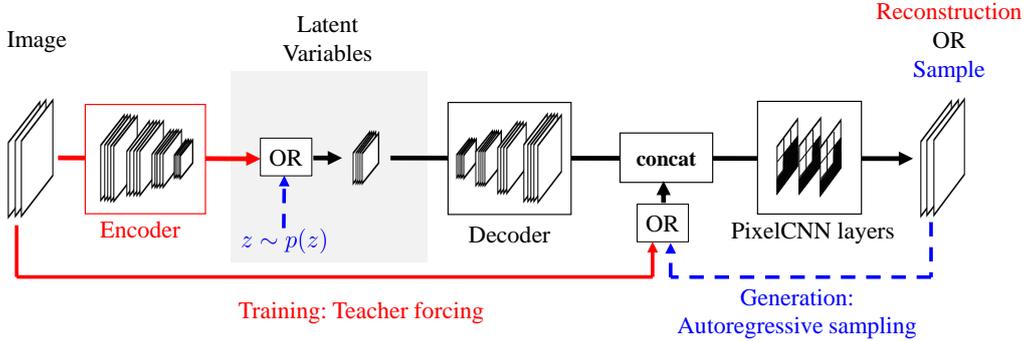

Figure 2: Our proposed model, PixelVAE, makes use of PixelCNN to model an autoregressive decoder for a VAE. VAEs, which assume (conditional) independence among pixels, are known to suffer from blurry samples, while PixelCNN, modeling the joint distribution, produces sharp samples, but lack a latent representation that might be more useful for downstream tasks. PixelVAE combines the best of both worlds, providing a meaningful latent representation, while producing sharp samples.

The idea of using autoregressive conditional likelihoods in VAEs has been explored in the context of sentence modeling in (Bowman et al., 2016), however in that work the use of latent variables fails to improve likelihood over a purely autoregressive model.

## 3 PixelVAE Model

Like a VAE, our model jointly trains an "encoder" inference network, which maps an image $x$ to a posterior distribution over latent variables $z$, and a "decoder" generative network, which models a distribution over $x$ conditioned on $z$. The encoder and decoder networks are composed of a series of convolutional layers, respectively with strided convolutions for downsampling in the encoder and transposed convolutions for upsampling in the decoder.

As opposed to most VAE decoders that model each dimension of the output independently (for example, by modeling the output as a Gaussian with diagonal covariance), we use a conditional PixelCNN in the decoder. Our decoder models $x$ as the product of each dimension $x_i$ conditioned on all previous dimensions and the latent variable $z$:

$$p(x|z) = \prod_i p(x_i|x_1, \ldots, x_{i-1}, z)$$

We first transform $z$ through a series of convolutional layers into feature maps with the same spatial resolution as the output image and then concatenate the resulting feature maps with the image. The resulting concatenated feature maps are then further processed by several PixelCNN masked convolutional layers and a final PixelCNN 256-way softmax output.

Unlike typical PixelCNN implementations, we use very few PixelCNN layers in our decoder, relying on the latent variables to model the structure of the input at scales larger than the combined receptive field of our PixelCNN layers. As a result of this, our architecture captures global structure at a much lower computational cost than a standard PixelCNN implementation.

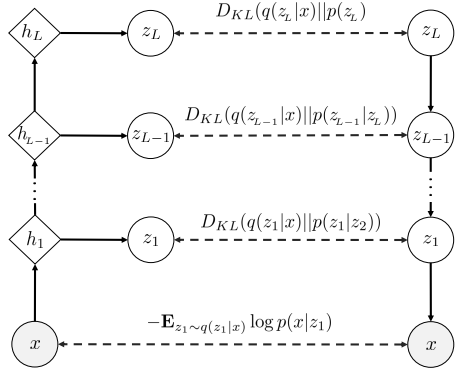

Figure 3: We generate top-down through a hierarchical latent space decomposition. The inference network generates latent variables by composing successive deterministic functions to compute parameters of the stochastic random variables. Dotted lines denote contributions to the cost.

### 3.1 HIERARCHICAL ARCHITECTURE

The performance of VAEs can be improved by stacking them to form a hierarchy of stochastic latent variables: in the simplest configuration, the VAE at each level models a distribution over the latent variables at the level below, with generation proceeding downward and inference upward through each level (i.e. as in Fig. 3). In convolutional architectures, the intermediate latent variables are typically organized into feature maps whose spatial resolution decreases toward higher levels.

Our model can be extended in the same way. At each level, the generator is a conditional PixelCNN over the latent features in the level below. This lets us autoregressively model not only the output distribution over pixels but also the prior over each set of latent feature maps. The higher-level PixelCNN decoders use diagonal Gaussian output layers instead of 256-way softmax, and model the dimensions within each spatial location (i.e. across feature maps) independently. This is done for simplicity, but is not a limitation of our model.

The output distributions over the latent variables for the generative and inference networks decompose as follows (see Fig. 3).

$$p(z_1, \cdots, z_L) = p(z_L)p(z_{L-1}|z_L) \cdots p(z_1|z_2)$$
$$q(z_1, \cdots, z_L|x) = q(z_1|x) \cdots q(z_L|x)$$

We optimize the negative of the evidence lower bound (sum of data negative log-likelihood and KL-divergence of the posterior over latents with the prior).

$$\begin{aligned}
-L(x,q,p) &= -E_{z_1 \sim q(z_1|x)} \log p(x|z_1) + D_{KL}(q(z_1, \cdots z_L|x)||p(z_1, \cdots, z_L)) \\
&= -E_{z_1 \sim q(z_1|x)} \log p(x|z_1) + \int_{z_1,\cdots,z_L} \prod_{j=1}^{L} q(z_j|x) \sum_{i=1}^{L} \log \frac{q(z_i|x)}{p(z_i|z_{i+1})} dz_1...dz_L \\
&= -E_{z_1 \sim q(z_1|x)} \log p(x|z_1) + \sum_{i=1}^{L} \int_{z_1,\cdots,z_L} \prod_{j=1}^{L} q(z_j|x) \log \frac{q(z_i|x)}{p(z_i|z_{i+1})} dz_1...dz_L \\
&= -E_{z_1 \sim q(z_1|x)} \log p(x|z_1) + \sum_{i=1}^{L} \int_{z_i,z_{i+1}} q(z_{i+1}|x)q(z_i|x) \log \frac{q(z_i|x)}{p(z_i|z_{i+1})} dz_i dz_{i+1} \\
&= -E_{z_1 \sim q(z_1|x)} \log p(x|z_1) + \sum_{i=1}^{L} \mathbf{E}_{z_{i+1} \sim q(z_{i+1}|x)} \big[ D_{KL}(q(z_i|x)||p(z_i|z_{i+1})) \big]
\end{aligned}$$

Note that by specifying an autoregressive PixelCNN prior over each latent level $z_i$, we can leverage masked convolutions (van den Oord et al., 2016b) and samples drawn independently from the approximate posterior $q(z_i \mid x)$ (i.e. from the inference network) to train efficiently in parallel on GPUs.

## 4 EXPERIMENTS

### 4.1 MNIST

| Model | NLL Test |
|---|---|
| DRAW (Gregor et al., 2016) | $\leq 80.97$ |
| Discrete VAE (Rolfe, 2016) | $= 81.01$ |
| IAF VAE (Kingma et al., 2016) | $\approx 79.88$ |
| PixelCNN (van den Oord et al., 2016a) | $= 81.30$ |
| PixelRNN (van den Oord et al., 2016a) | $= 79.20$ |
| Convolutional VAE | $\leq 87.41$ |
| PixelVAE | $\leq 80.64$ |
| Gated PixelCNN (our implementation) | $= 80.10$ |
| Gated PixelVAE | $\approx 79.48\ (\leq 80.02)$ |
| Gated PixelVAE without upsampling | $\approx \mathbf{79.02}\ (\leq 79.66)$ |

Table 1: We compare performance of different models on binarized MNIST. "PixelCNN" is the model described in van den Oord et al. (2016a). Our corresponding latent variable model is "Pixel-VAE". "Gated PixelCNN" and "Gated PixelVAE" use the gated activation function in van den Oord et al. (2016b). In "Gated PixelVAE without upsampling", a linear transformation of latent variable conditions the (gated) activation in every PixelCNN layer instead of using upsampling layers.

We evaluate our model on the binarized MNIST dataset (Salakhutdinov & Murray, 2008; Lecun et al., 1998) and report results in Table 1. We also experiment with a variant of our model in which each PixelCNN layer is directly conditioned on a linear transformation of latent variable, $z$ (rather than transforming $z$ first through several upsampling convolutional layers) (as in (van den Oord et al., 2016b) and find that this further improves performance, achieving an NLL upper bound comparable with the current state of the art. We estimate the marginal NLL of our model (using 1000 importance samples per datapoint) and find that it achieves state-of-the-art performance.

#### 4.1.1 NUMBER OF PIXELCNN LAYERS

The masked convolutional layers in PixelCNN are computationally expensive because they operate at the full resolution of the image and in order to cover the full receptive field of the image, PixelCNN typically needs a large number of them. One advantage of our architecture is that we can achieve strong performance with very few PixelCNN layers, which makes training and sampling from our model significantly faster than PixelCNN. To demonstrate this, we compare the performance of our model to PixelCNN as a function of the number of PixelCNN layers (Fig. 4a). We find that with fewer than 10 autoregressive layers, our PixelVAE model performs much better than PixelCNN. This is expected since with few layers, the effective receptive field of the PixelCNN output units is too small to capture long-range dependencies in the data.

We also observe that adding even a single PixelCNN layer has a dramatic impact on the NLL bound of PixelVAE. This is not surprising since the PixelCNN layer helps model local characteristics which are complementary to the global characteristics which a VAE with a factorized output distribution models.

In our MNIST experiments, we have used PixelCNN layers with no blind spots using vertical and horizontal stacks of convolutions as proposed in (van den Oord et al., 2016b).

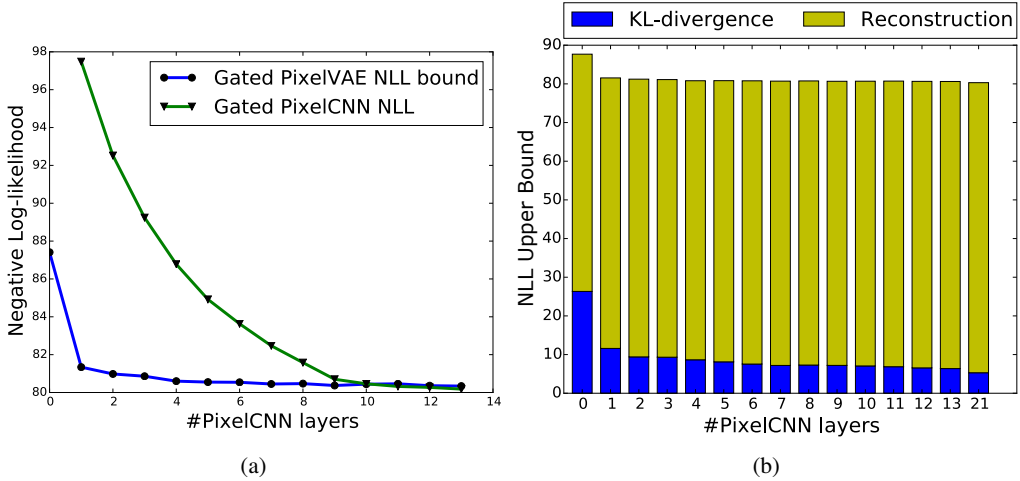

Figure 4: (a) Comparison of Negative log-likelihood upper bound of PixelVAE and NLL for Pixel-CNN as a function of the number of PixelCNN layers used. (b) Cost break down into KL divergence and reconstruction cost.

#### 4.1.2 LATENT VARIABLE INFORMATION CONTENT

Because the autoregressive conditional likelihood function of PixelVAE is expressive enough to model some properties of the image distribution, it isn't forced to account for those properties through its latent variables as a standard VAE is. As a result, we can expect PixelVAE to learn latent representations which are invariant to textures, precise positions, and other attributes which are more efficiently modeled by the autoregressive decoder. To empirically validate this, we train PixelVAE models with different numbers of autoregressive layers (and hence, different PixelCNN receptive field sizes) and plot the breakdown of the NLL bound for each of these models into the reconstruction term $\log p(x|z)$ and the KL divergence term $D_{KL}(q(z|x)||p(z))$ (Fig. 4b). The KL divergence term can be interpreted as a measure of the information content in the posterior distribution $q(z|x)$ and hence, models with smaller KL terms encode less information in their latent variables.

We observe a sharp drop in the KL divergence term when we use a single autoregressive layer compared to no autoregressive layers, indicating that the latent variables have been freed from having to encode small-scale details in the images. Since the addition of a single PixelCNN layer allows the decoder to model interactions between pixels which are at most 2 pixels away from each other (since our masked convolution filter size is $5 \times 5$), we can also say that most of the non-trivial (long-range) structure in the images is still encoded in the latent variables.

### 4.2 LSUN BEDROOMS

To evaluate our model's performance with more data and complicated image distributions, we perform experiments on the LSUN bedrooms dataset (Yu et al., 2015). We use the same preprocessing as in Radford et al. (2015) to remove duplicate images in the dataset. For quantitative experiments we use a $32 \times 32$ downsampled version of the dataset, and we present samples from a model trained on the $64 \times 64$ version.

We train a two-level PixelVAE with latent variables at $1 \times 1$ and $8 \times 8$ spatial resolutions. We find that this outperforms both a two-level convolutional VAE with diagonal Gaussian output and a single-level PixelVAE in terms of log-likelihood and sample quality. We also try replacing the PixelCNN layers at the higher level with a diagonal Gaussian decoder and find that this hurts log-likelihood, which suggests that multi-scale PixelVAE uses those layers effectively to autoregressively model latent features.

### 4.2.1 FEATURES MODELED AT EACH LAYER

To see which features are modeled by each of the multiple layers, we draw multiple samples while varying the sampling noise at only a specific layer (either at the pixel-wise output or one of the latent layers) and visually inspect the resulting images (Fig. 5). When we vary only the pixel-level sampling (holding $z_1$ and $z_2$ fixed), samples are almost indistinguishable and differ only in precise positioning and shading details, suggesting that the model uses the pixel-level autoregressive distribution to model only these features. Samples where only the noise in the middle-level ($8 \times 8$) latent variables is varied have different objects and colors, but appear to have similar basic room geometry and composition. Finally, samples with varied top-level latent variables have diverse room geometry.

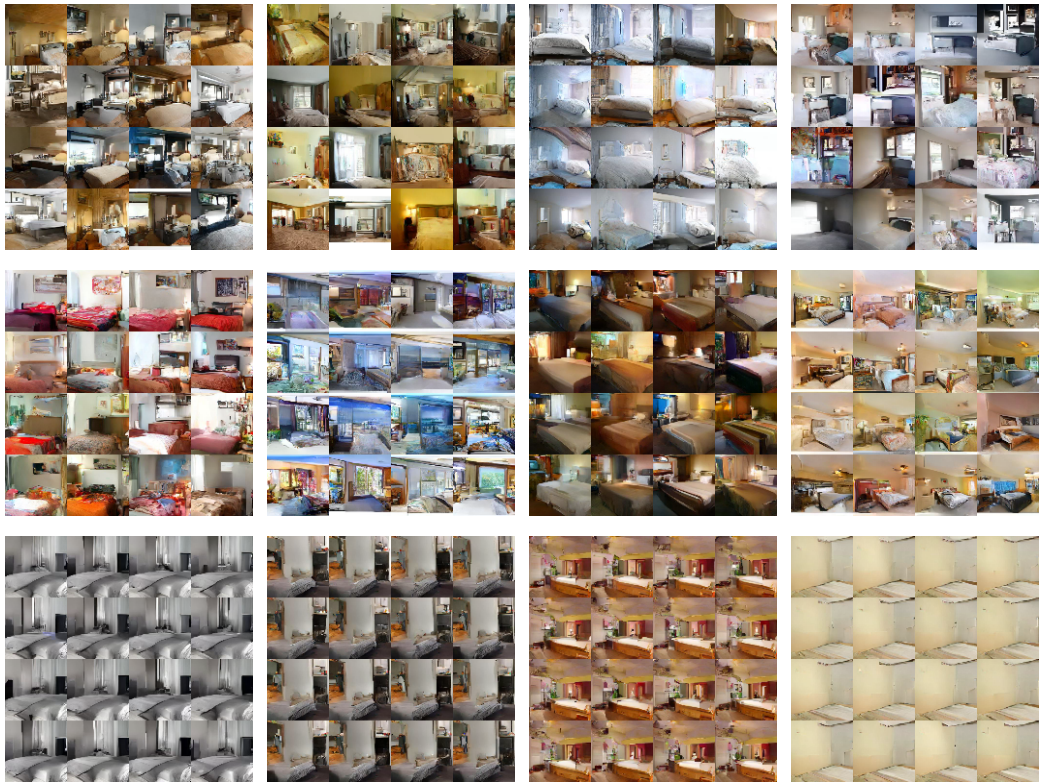

Figure 5: We visually inspect the variation in image features captured by the different levels of stochasticity in our model. For the two-level latent variable model trained on $64 \times 64$ LSUN bedrooms, we vary only the top-level sampling noise *(top)* while holding the other levels constant, vary only the middle-level noise *(middle)*, and vary only the bottom (pixel-level) noise *(bottom)*. It appears that the top-level latent variables learn to model room structure and overall geometry, the middle-level latents model color and texture features, and the pixel-level distribution models low-level image characteristics such as texture, alignment, shading.

## 4.3 $64 \times 64$ IMAGENET

The $64 \times 64$ ImageNet generative modeling task was introduced in (van den Oord et al., 2016a) and involves density estimation of a difficult, highly varied image distribution. We trained a heirarchical PixelVAE model (with a similar architecture to the model in section 4.2) of comparable size to the models in van den Oord et al. (2016a;b) on $64 \times 64$ ImageNet in 5 days on 3 NVIDIA GeForce GTX 1080 GPUs. We report validation set likelihood in Table 2. Our model achieves a slightly lower log-likelihood than PixelRNN (van den Oord et al., 2016a), but a visual inspection of ImageNet samples from our model (Fig. 6) reveals them to be significantly more globally coherent than samples from PixelRNN.

| Model | NLL Validation (Train) |
| --- | --- |
| Convolutional DRAW (Gregor et al., 2016) | $\leq$ 4.10 (4.04) |
| Real NVP (Dinh et al., 2016) | $=$ 4.01 (3.93) |
| PixelRNN (van den Oord et al., 2016a) | $=$ 3.63 (3.57) |
| Gated PixelCNN (van den Oord et al., 2016b) | $=$ **3.57** (3.48) |
| Hierarchical PixelVAE | $\leq$ 3.66 (3.59) |

Table 2: Model performance on 64x64 ImageNet.

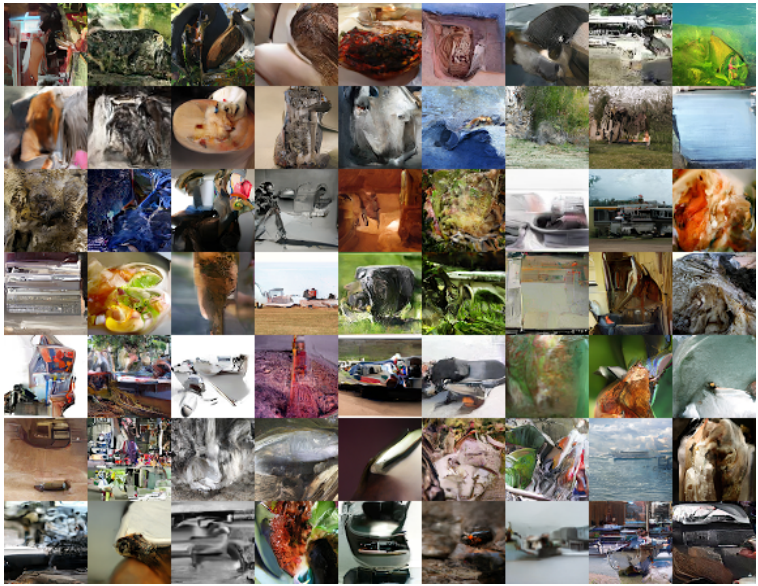

Figure 6: Samples from hierarchical PixelVAE on the 64x64 ImageNet dataset.

## 5 CONCLUSIONS

In this paper, we introduced a VAE model for natural images with an autoregressive decoder that achieves strong performance across a number of datasets. We explored properties of our model, showing that it can generate more compressed latent representations than a standard VAE and that it can use fewer autoregressive layers than PixelCNN. We established a new state-of-the-art on binarized MNIST dataset in terms of likelihood on $64 \times 64$ ImageNet and demonstrated that our model generates high-quality samples on LSUN bedrooms.

The ability of PixelVAE to learn compressed representations in its latent variables by ignoring the small-scale structure in images is potentially very useful for downstream tasks. It would be interesting to further explore our model's capabilities for semi-supervised classification and representation learning in future work.

### ACKNOWLEDGMENTS

The authors would like to thank the developers of Theano (Theano Development Team, 2016) and Blocks and Fuel (van Merriënboer et al., 2015). We acknowledge the support of the following agencies for research funding and computing support: Ubisoft, Nuance Foundation, NSERC, Calcul Quebec, Compute Canada, CIFAR, MEC Project TRA2014-57088-C2-1-R, SGR project 2014-SGR-1506 and TECNIOspring-FP7-ACCI grant.